\documentclass[journal]{IEEEtran}

\usepackage[utf8]{inputenc}
\usepackage[T1]{fontenc}
\usepackage{amsmath,amssymb,amsfonts}
\usepackage{amsthm}
\usepackage{graphicx}
\usepackage{booktabs}
\usepackage{multirow}
\usepackage{cite}
\usepackage{url}
\usepackage[hidelinks]{hyperref}
\usepackage{xcolor}

\newtheorem{definition}{Definition}

\title{SafeAgent: A Runtime Protection Architecture for Agentic Systems}

\author{
Hailin Liu, Eugene Ilyushin, Jie Ni, and Min Zhu%
\thanks{Hailin Liu, Lomonosov Moscow State University, e-mail: hereishailin@outlook.com.}%
\thanks{Eugene Ilyushin, Lomonosov Moscow State University, Central University, e-mail:eugene.ilyushin@gmail.com.}%
\thanks{Jie Ni, Lomonosov Moscow State University, e-mail: nijie4207@ gmail.com.}%
\thanks{Min Zhu, Lomonosov Moscow State University, e-mail: zhumingdehao@gmail.com.}%
}

\begin{document}

\maketitle

\begin{abstract}
Large language model (LLM) agents are vulnerable to prompt-injection attacks that propagate through multi-step workflows, tool interactions, and persistent context, making input-output filtering alone insufficient for reliable protection. This paper presents SafeAgent, a runtime security architecture that treats agent safety as a stateful decision problem over evolving interaction trajectories. The proposed design separates execution governance from semantic risk reasoning through two coordinated components: a runtime controller that mediates actions around the agent loop and a context-aware decision core that operates over persistent session state. The core is formalized as a context-aware advanced machine intelligence and instantiated through operators for risk encoding, utility-cost evaluation, consequence modeling, policy arbitration, and state synchronization. Experiments on Agent Security Bench (ASB) and InjecAgent show that SafeAgent consistently improves robustness over baseline and text-level guardrail methods while maintaining competitive benign-task performance. Ablation studies further show that recovery confidence and policy weighting determine distinct safety–utility operating points.
\end{abstract}



\section{Introduction}
\label{sec:intro}

Large language models (LLMs) are increasingly deployed as agentic systems that not only generate text, but also plan actions, invoke tools, and interact with external environments. In such settings, the model no longer operates as an isolated conversational component: it becomes part of a workflow that reads external observations, updates context, selects actions, and produces side effects through tool execution. ReAct-style agents are a representative example of this paradigm, showing how reasoning and acting can be interleaved within a single execution loop \cite{yao2023react}. While this design significantly expands agent capability, it also turns security failures into operational failures.

Among these threats, prompt injection has emerged as a central challenge for tool-using agents. In agentic workflows, malicious instructions may enter not only through direct user inputs, but also through retrieved documents, tool outputs, memory, or other external artifacts later incorporated into the working context \cite{greshake2023indirectpromptinjection}. As a result, attacks are no longer limited to single-turn prompt manipulation: they may unfold incrementally across the reasoning--action loop, influence intermediate planning, and ultimately induce unsafe tool use or unauthorized data transfer.

Existing defenses only partially address this problem. Many practical safeguards still operate at the level of local inputs and outputs, which makes them difficult to apply to multi-step workflows where risk propagates through observations, tool interactions, and accumulated context. More structured system-level solutions exist, but they are often tied to particular agent designs or execution assumptions. What remains missing is a general runtime security architecture that can mediate execution while making decisions over persistent context state.

To address this gap, we introduce a new runtime protection architecture for agentic systems. The proposed design separates execution governance from context-aware security reasoning: one layer mediates execution around the agent loop and privileged interactions, while another performs stateful risk assessment and intervention over persistent session context. Rather than treating safety as a single-turn filtering problem, this architecture frames agent protection as runtime decision-making over evolving interaction trajectories.

The main contributions of this work are as follows:
\begin{itemize}
    \item We formulate agent security as a \emph{runtime} and \emph{stateful} decision problem, rather than as isolated input-output filtering.
    \item We propose a runtime protection architecture for agentic systems that separates execution governance from context-aware security reasoning.
    \item We formalize the decision component of this architecture as a context-aware AMI system and instantiate it through operators for risk encoding, utility--cost evaluation, consequence modeling, policy arbitration, and state synchronization.
    \item We evaluate the proposed architecture on two representative agent-security benchmarks, ASB and InjecAgent, and demonstrate stronger robustness than baseline and text-level guardrail methods.
    \item We further analyze the effects of recovery confidence and policy weighting through ablation studies, showing that both parameters significantly affect the safety--utility operating point of the system.
\end{itemize}

The rest of the paper is organized as follows. Section~II introduces the background and threat setting. Section~III reviews related work. Sections~IV and~V present the proposed runtime controller and decision core, respectively. Section~VI reports the experimental evaluation, and Section~VII concludes the paper.

\section{Background}
\label{sec:background}

\subsection{Prompt Injection}
\label{subsec:prompt_injection}

Prompt injection is an adversarial technique that manipulates the instructions a language model conditions on, causing it to deviate from the developer’s intended policy or task objective. In agent-based systems, the fundamental problem arises when untrusted text (e.g., retrieved documents, web pages, tool outputs) is implicitly treated as control input rather than data. This collapses the boundary between content and instruction and creates a systemic attack surface whenever heterogeneous context is concatenated and delegated to the model.

Prompt injection can be categorized as \emph{direct} or \emph{indirect}. Direct injection is carried by user input and explicitly attempts to override system constraints (e.g., “ignore all previous instructions”). Indirect injection is embedded in third-party artifacts later consumed by the agent. Greshake et al.~\cite{greshake2023indirectpromptinjection} show that such indirect attacks can compromise real-world LLM-integrated applications, especially when malicious instructions embedded in external content are elevated into the model’s execution context. 

Recent work emphasizes that prompt injection is not limited to single-turn manipulation but emerges at the \emph{workflow level}. Ferrag et al.~\cite{ferrag2025workflowthreats} analyze how multi-stage agent pipelines (retrieval $\rightarrow$ planning $\rightarrow$ tool execution $\rightarrow$ memory update) enable composite attacks, where injected payloads are distributed across steps to achieve escalation or persistence. In addition, instruction-triggered backdoor behaviors in customized models~\cite{zhang2024instruction} further expand the threat model: even in the absence of explicit override strings, specific patterns may activate undesirable plans or unsafe disclosures.

\subsection{ReAct Agent}
\label{subsec:react-agent}

ReAct (Reason+Act) is a prompting paradigm that interleaves explicit reasoning steps with external actions and conditions subsequent reasoning on returned observations. Instead of generating a final answer in a single pass, a ReAct agent iterates over a loop of reasoning, action selection, and observation grounding. Yao et al.~\cite{yao2023react} demonstrate that this interleaving improves task performance in interactive and multi-hop settings.

\begin{figure}[!h]
\centering
\includegraphics[width=\linewidth]{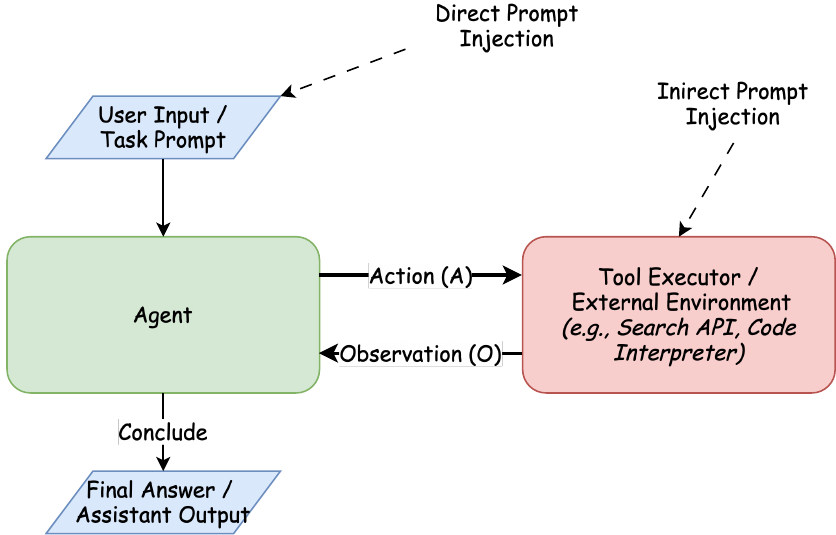}
\caption{Operation of a ReAct agent and representative attack surfaces in the action-observation loop.}
\label{fig:react_baseline}
\end{figure}

While the action--observation loop enhances capability, it also expands the attack surface. Observations (tool outputs or retrieved content) are repeatedly incorporated into the reasoning context, allowing adversarial strings embedded in external environments to influence subsequent decisions. When agents execute high-impact actions based on contaminated observations, prompt injection becomes operational rather than purely textual.

\subsection{Advanced Machine Intelligence}
\label{subsec:ami}

We use the term \emph{Advanced Machine Intelligence (AMI)} to describe a system-level architecture for autonomous agents that integrate perception, memory, prediction, and decision-making over extended horizons. A canonical decomposition includes: perception modules that encode observations into latent representations; a world model that predicts future latent states conditioned on candidate actions; a policy that proposes actions; an evaluation component that scores predicted trajectories; and memory mechanisms operating across multiple time scales~\cite{lecun2022path}.

The world model functions as an internal simulator that enables counterfactual rollouts. Candidate actions are propagated through the predictive model to generate hypothetical future states, which are then evaluated before execution. This predictive–evaluative loop supports planning under partial observability and reduces reliance on purely reactive behavior. Furthermore, the separation between short-term state and longer-term memory provides stability against transient noise while allowing adaptation to persistent environmental signals.

\section{Related Work}
\label{sec:related_work}

\subsection{Model-level robustness}
\label{subsec:model_robustness}

A natural line of defense is to improve the model’s intrinsic resistance to instruction hijacking through training-time interventions. JATMO \cite{piet2024jatmo} represents this direction by optimizing a customized model to better preserve user-task adherence under prompt injection, and by stress-testing the resulting robustness under stronger, search-based attacks. In practice, such training-time hardening can reduce the probability that malicious instructions override the intended objective, but it does not provide structural guarantees once high-privilege actions are available.

More recent work emphasizes reducing deployment-time overhead while maintaining benign utility. RedVisor \cite{liu2026redvisor} decouples \emph{inspection} from \emph{response} inside a single serving pipeline by enabling a lightweight, removable adapter only in a reasoning phase, and then muting it for response generation. By reusing the inspection KV cache for generation, RedVisor aims to avoid the doubled prefill cost of two-model guardrail pipelines while still making injection localization and rejection conditions explicit.

Complementary to training and serving optimizations, long-context settings motivate fine-grained sanitization mechanisms. PISanitizer \cite{geng2025pisanitizer} performs targeted removal of the most influential instruction-like spans (rather than blocking entire samples), which is especially relevant when injected content occupies only a small fraction of a large retrieved context. This token/span-level perspective also sets up the next section on detection and filtering systems.

Model-level improvements can lower the likelihood of hijacking, but they cannot enforce system-level least-privilege or isolation guarantees; therefore they must be combined with architecture- and execution-level defenses.

\subsection{Detection-based Guardrails and Filters}
\label{subsec:guardrails_filters}

A dominant line of defense since 2023 has been the deployment of guardrails and filtering mechanisms around LLM. Instead of modifying the base model, these approaches insert intermediate layers that analyze inputs, outputs, or intermediate reasoning traces in order to detect and suppress prompt injection attempts. Such mechanisms rapidly gained popularity in both academic prototypes and commercial products because they are modular, API-compatible, and relatively easy to integrate into existing agent pipelines.

Early guardrail systems focused on semantic classification and policy filtering. Llama Guard~\cite{inan2023llamaguard} exemplifies this direction by using a dedicated LLM-based safeguard model to screen inputs and outputs against predefined safety policies. Similarly, toolkits such as LLM Guard~\cite{protectai_llmguard} package multiple scanners into configurable pipelines, enabling layered inspection of user queries and model responses. These systems reflect an engineering paradigm in which safety is implemented as a middleware layer, separating detection logic from the core reasoning model.

Subsequent work explored stronger detection signals that leverage internal model sensitivities. GradSafe~\cite{xiesafetygrad2024} analyzes safety-critical gradients to identify adversarial prompts, demonstrating improved robustness against adaptive jailbreak-style attacks, albeit at higher computational cost. GuardReasoner~\cite{liu2025guardreasoner} further advances reasoning-based safeguards by generating structured safety judgments rather than relying solely on surface-level features. However, increasing guardrail strength introduces a new challenge: excessive conservatism. InjecGuard~\cite{li2024injecguard} systematically studies the phenomenon of \emph{over-defense}, where aggressive filtering reduces model utility and blocks benign instructions, highlighting an inherent safety--utility trade-off.

Although detection-based guardrails have become widely deployed and commercially mature, their limitations have gradually emerged. As attacks evolve from simple override strings to multi-stage workflow manipulation, gradient-based optimization, and long-context contamination, purely detection-oriented strategies struggle to provide structural guarantees. This shift has motivated a growing body of research that moves beyond filtering and toward system-level architectural constraints.

\subsection{System-level defenses}
\label{subsec:system_defenses}

\begin{figure*}[!t]
\includegraphics[width=\textwidth]{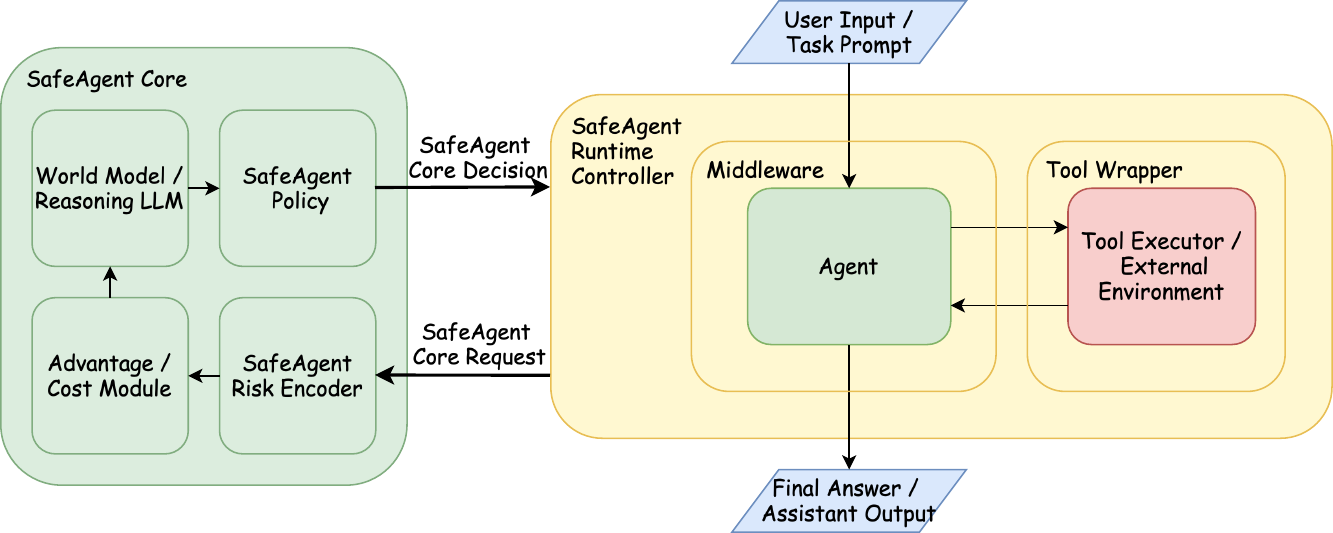}
\centering
\caption{Overview of the SafeAgent architecture.}
\label{fig:safeagent_system_overview}
\end{figure*}

A common position in practical agent security is that prompt injection cannot be removed solely by making the model more obedient. Instead, the system should limit the \emph{blast radius} of any compromised step: apply least privilege, complete mediation, and fail-safe defaults so that untrusted text cannot directly trigger high-impact side effects.

A complementary direction makes \emph{instruction hierarchy} an explicit control rule. Rather than relying on prompt phrasing, the system enforces a priority order over instruction sources (e.g., system/developer constraints dominate user requests, while tool outputs remain untrusted). Wallace et al.~\cite{wallace2024instructionhierarchy} formalize this hierarchy and show that training can improve consistency under instruction conflicts. In agent workflows, hierarchy is most effective when paired with structured prompts and runtime checks that preserve the boundary between \emph{control} and \emph{data} across turns.

Beyond hierarchy, system designs increasingly use \emph{isolation} to prevent untrusted content from shaping privileged decisions. Wen et al.~\cite{wen2026agentsys} propose hierarchical memory management where a main agent delegates tool-facing subtasks to isolated workers, and only schema-validated structured results are allowed to cross boundaries, reducing persistence of injected instructions.

Another step is to constrain execution via explicit \emph{graphs}. Instead of letting the model synthesize arbitrary tool calls, the system executes within a tool-dependency structure, turning tool choice into constrained traversal. An et al.~\cite{an2025ipiguard} instantiate this idea with a Tool Dependency Graph that blocks out-of-graph invocations, addressing the common failure mode where injected text introduces new tools or steers parameters toward exfiltration.

These principles appear in production-oriented guardrail frameworks focused on configurability and deployability. NeMo Guardrails~\cite{rebuffi2023nemoguardrails} provides a programmable control layer (rails) to specify allowed conversational flows, tool usage patterns, and safety checks around an LLM. However, system-level defenses are often \emph{task- and integration-specific}: they depend on correctly identifying privileged actions, defining boundaries, and maintaining policies as tools and workflows evolve.

\subsection{Insights and Motivation}
\label{subsec:insights_motivation}

Across prior work, prompt injection defenses can be grouped into three complementary directions: model-level robustness improves a model’s intrinsic resistance to instruction hijacking but cannot enforce least privilege once high-impact capabilities are available; detection-based guardrails and filters add modular monitoring around inputs, outputs, and intermediate artifacts yet face recurring safety--utility and latency--coverage trade-offs under adaptive, workflow-level attacks; and system-level defenses reduce reliance on model obedience by enforcing structural constraints such as instruction-source hierarchy, isolation boundaries, and constrained execution graphs to limit blast radius and prevent untrusted content from steering privileged actions. Taken together, these directions motivate a stateful view of agent security in which risk is not a single-turn property but accumulates and propagates through the action--observation loop.

A stateful perspective naturally leads to the emerging idea of \emph{taint tracking}\cite{greshake2023indirectpromptinjection} for LLM agents: treat untrusted content as carrying a contamination label and propagate it through the agent’s intermediate representations and decision points. Unlike sample-level filtering, taint tracking focuses on \emph{information flow} across a workflow: external observations (e.g., retrieval snippets, tool outputs, logs) can influence planning, tool arguments, and memory updates, turning a localized injection into persistent control over future behavior. Recent system-oriented proposals increasingly converge on the same principle: safety decisions should be conditioned on where information originated, how it was transformed, and whether it is about to cross a privileged boundary (e.g., invoking a high-impact tool, writing long-term memory, or emitting sensitive content).

This motivates a runtime architecture in which contamination is represented as explicit state and checked at critical execution boundaries.


\section{SafeAgent Runtime Controller: Governed Interface and Context Manager}
\label{sec:safeagent_controller}

The SafeAgent Runtime Controller is designed as a governed interface between an LLM agent and its execution environment. Rather than embedding safety logic directly into application code or relying on the model to consistently follow instruction hierarchies, the Controller centralizes request handling, context management, and operational governance in a single middleware layer. This separation enables the agent to remain task-focused while the Controller enforces execution discipline and provides a uniform surface for policy application, auditing, and recovery.

At a high level, the Controller wraps the agent loop and manages information and action flow across a session. It curates working context (user inputs, tool observations, intermediate artifacts, and session metadata) and issues structured requests to an external decision module (SafeAgent Core) when risk assessment or action arbitration is needed. Given the current observation and relevant context, the Core returns a decision signal with optional guidance; the Controller enforces this decision while keeping privileged interfaces insulated from untrusted content.

A central objective is to enforce controlled execution boundaries for tools and external environments. By routing all tool use through a governed wrapper, the Controller applies consistent operational policies across heterogeneous backends (e.g., shell, web, databases, third-party APIs), limiting the ability of adversarial observations to induce high-impact side effects.

The Controller also provides resilience for long-running sessions. It handles transient failures via retries, triggers safe replanning when execution diverges from the session objective, and contains tool-level faults to avoid cascading errors. When constraints are violated or operational budgets are exceeded, it either delays execution until resources become available or blocks the risky step under the configured governance mode.

The rest of this section details the Controller design in terms of session-state management, governed tool mediation, and recovery-oriented orchestration around the agent loop.

\subsection{Session Context as Runtime Object}
\label{subsec:controller_context}

In agentic workflows, the operative context is not a single-turn prompt but a trajectory formed by interleaved user inputs, model outputs, tool invocations, and environment feedback. Each step can change the feasible action space and reshape subsequent planning, turning the workflow into a dynamically evolving control process rather than a static conversation.

We represent session context using two complementary layers, a trace that records key, auditable events produced during an interaction; and a runtime snapshot that serves as the Controller's working memory for real-time control, enabling fast decisions and interventions.

The Controller normalizes each interaction step into a structured request to the Core that contains the necessary observation signal for downstream decision-making. After receiving a decision from the Core, the Controller applies the decision to intervene in or merge the current context snapshot, and to route execution across the agent graph according to the prescribed control flow.

\begin{figure}[!h]
\centering
\includegraphics[width=\linewidth]{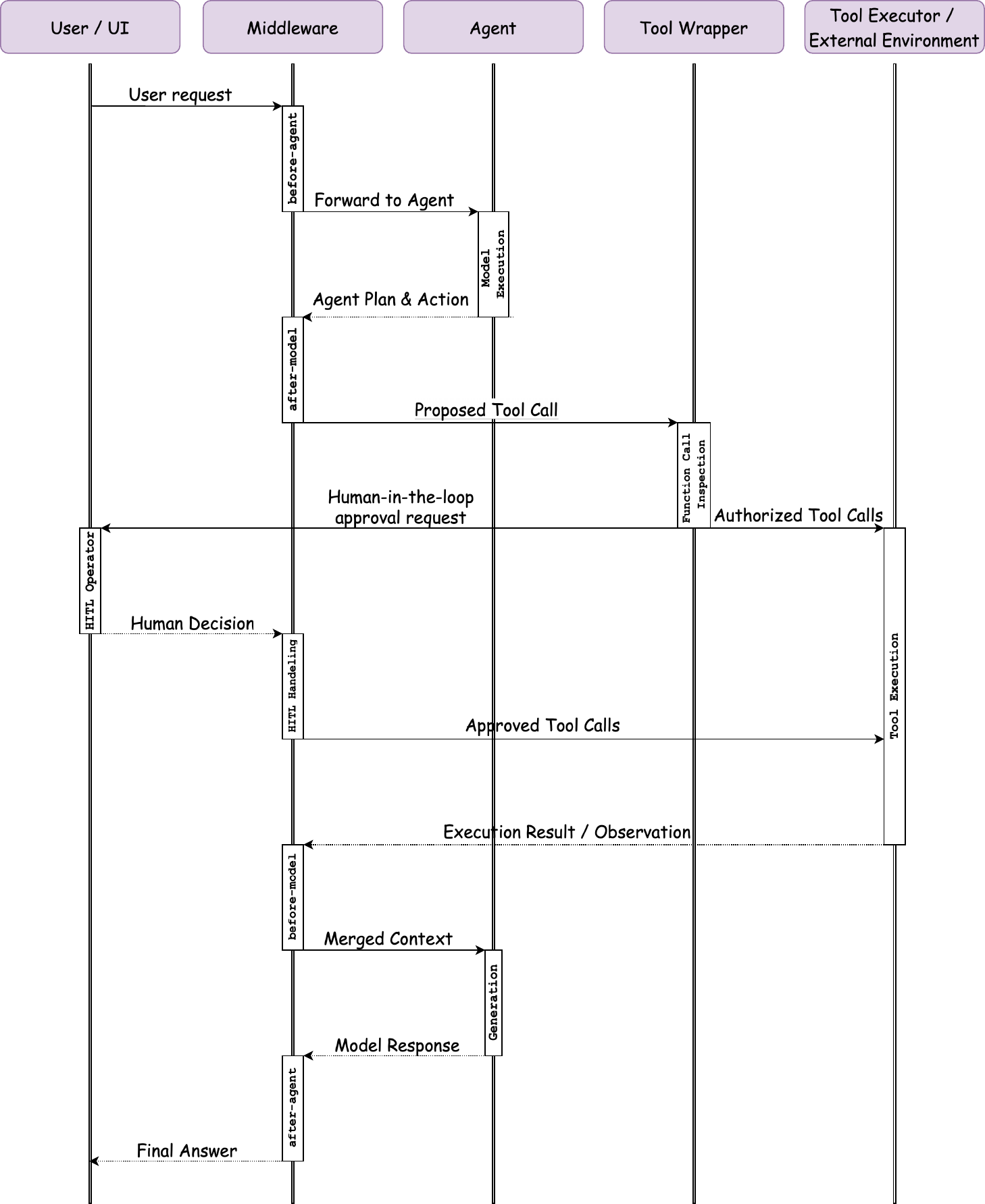}
\caption{Flow control sequence in the agent lifecycle.}
\label{fig:controller_sequence}
\end{figure}

Figure~\ref{fig:controller_sequence} illustrates the Controller’s view of a single step as a phase-structured lifecycle from request intake to final response. We abstract the runtime inspection process as a set of middleware components with four complementary roles: user input inspection, model plan inspection, observation inspection, and model output inspection. These components respectively govern incoming user requests, intermediate action intentions, external observations from tools or environment, and the final assistant response.

At each hook, the control plane may allow the step to proceed, block the step to prevent unsafe progression, or attempt to repair a degraded context. By anchoring interventions to explicit phase boundaries and operating over merged context snapshots rather than raw text concatenation, the Controller supports systematic review, targeted correction, and post-hoc audit.

\subsection{Governed Tool Access through a Controlled Boundary}
\label{subsec:controller_tools}

In agentic workflows, tool invocation is not merely text generation but a privileged operation that can induce external side effects. Consequently, the impact of prompt injection is amplified in tool-using settings, since injected instructions may be directly translated into filesystem, network, or API actions.

Within the Controller, the \emph{tool wrapper} acts as a governed execution boundary between the agent and the external environment. Instead of executing tool calls directly, proposed invocations are intercepted, normalized into structured, auditable requests, and subjected to inspection before any side effects occur. Each invocation is represented with explicit metadata—including tool name, arguments, and declared capability—enabling consistent and context-aware governance across heterogeneous tools without relying on raw text interpretation.

Authorization decisions are conditioned on accumulated workflow-level contamination signals rather than the current invocation alone. Based on the agent's proposed plan, the developer-provided tool specification, and the evolving session context, the Controller can allow execution, block the call, rewrite unsafe arguments, or escalate the request for human-in-the-loop approval. In addition, the framework supports operational rate governance, so tool execution is constrained not only by risk-based arbitration but also by configurable call-budget availability.

\subsection{Recovery-Oriented Orchestration around the Agent Loop}
\label{subsec:controller_recovery}

Beyond static mediation, the SafeAgent Controller implements a set of recovery-oriented orchestration strategies around the agent loop. These strategies allow the runtime to repair corrupted context, retry or redirect unsafe plans, and safely stop execution when necessary.

\paragraph{Context override}
The Controller supports recovery-oriented execution by allowing unsafe content to be selectively rewritten at inspection points. User inputs, agent responses, and external observations can be sanitized before being propagated, preventing unsafe signals from influencing subsequent reasoning while preserving task continuity.

\paragraph{Agent replanning}
Replanning provides a recovery mechanism at the decision level, complementing content-level overrides. When the agent's proposed action becomes misaligned with the task objective or exhibits elevated risk, the Controller triggers a constrained replan step, prompting the model to reconsider and revise its next action.

\paragraph{Context rollback}
When external observations are identified as unsafe or misleading, the Controller treats the current execution trajectory as unreliable and rolls back the runtime state to a prior trusted checkpoint. Execution then resumes from this restored state, allowing the agent to follow an alternative trajectory without inheriting the effects of the compromised observation.

\paragraph{Session termination and safe mode}
When risks cannot be effectively mitigated through local recovery, the Controller provides a fail-safe mechanism by terminating the current session or enforcing a safe mode. This ensures that execution does not proceed once critical safety constraints are violated.

\paragraph{Tool argument rewriting}
The Controller enables partial recovery at the action level by rewriting tool invocation parameters under safety constraints. This constrains the operational scope of actions without blocking execution, preserving task utility while enforcing capability-aligned behavior.

\paragraph{Human approval and shadow execution}
For actions requiring stronger guarantees, the Controller can escalate tool invocations to human-in-the-loop approval or developer-provided shadow execution. Such mechanisms are essential in safety-critical systems, serving as a final safeguard when automated control is insufficient and ensuring that high-impact operations are explicitly validated before execution.

Overall, these recovery mechanisms establish the Controller as an active runtime governor that maintains execution integrity under dynamic and potentially adversarial conditions. By supporting context repair, decision revision, trajectory rollback, constrained action execution, and fail-safe termination, SafeAgent enables resilient agent operation in the presence of context contamination, tool failures, and injection-induced deviations.

The Runtime Controller thus defines the governed execution interface of SafeAgent, managing context propagation and mediating interactions with external environments. However, it does not perform semantic risk reasoning or action-level arbitration. Instead, these responsibilities are delegated to a dedicated decision module that evaluates observations under accumulated context and produces structured safety signals. The next section introduces the design of SafeAgent Core, which implements context-aware risk assessment and decision-making.


\section{SafeAgent Core: A Context-Aware Decision Core for Risk Assessment and Intervention}
\label{sec:safeagent_core}

While the Runtime Controller governs execution and mediates interactions with the environment, risk-sensitive decisions are delegated to an external component: the \emph{SafeAgent Core}. The Core serves as a context-aware decision module that evaluates the safety of agent behavior under accumulated session state and produces structured intervention signals.

The Core operates purely at the semantic level. Given the current observation and the maintained context state, it does not execute actions or interact with external systems, but instead determines whether the ongoing workflow should proceed, be modified, or be halted. This design isolates decision-making from execution, enabling flexible and policy-driven control over agent behavior.

At each inspection point, the Controller submits a normalized request consisting of the current observation and the aggregated context. The Core maintains a persistent internal state across steps, capturing risk-relevant information such as historical signals, inferred intent, and prior interventions. Based on this state, it produces a structured decision output that guides subsequent execution.

This separation establishes SafeAgent as a two-level system: the Controller enforces execution constraints, while the Core performs stateful risk assessment and decision-making. In the remainder of this section, we describe the Core architecture in detail, focusing on its mechanisms for context-aware reasoning, action evaluation, and intervention generation.

\begin{definition}[Context-Aware Advanced Machine Intelligence]
\label{defin:core_formalization}
A context-aware Advanced Machine Intelligence (AMI) is a stateful decision system operating over partially observable interaction streams through latent representation and predictive modeling.

Let $\mathcal{X}$ denote the observation space, $\mathcal{Z}$ the latent representation space, $\mathcal{S}$ the internal state space, and $\mathcal{A}$ the action space. At each step $t$, the system maps the observation $x_t \in \mathcal{X}$ to a latent representation
\[
z_t = \Phi(x_t, s_t),
\]
and maintains an internal state $s_t \in \mathcal{S}$ that summarizes the interaction history.

The system includes a predictive model
\[
\hat{s}_{t+1} = \Psi(s_t, z_t, a),
\]
which estimates the evolution of the state under candidate actions.

Decision-making is performed by selecting an action
\[
a_t = \Pi(s_t, z_t, \hat{s}_{t+1}),
\]
based on the current state, latent representation, and predicted outcomes.

The internal state evolves according to
\[
s_{t+1} = \mathcal{M}(s_t, z_t, a_t),
\]
enabling context-aware reasoning across the interaction trajectory.
\end{definition}

Such a system performs decision-making through latent-space prediction and consequence-aware reasoning, rather than direct reactive mapping from observations to actions.

\begin{figure*}[t]
  \centering
  \includegraphics[width=\textwidth]{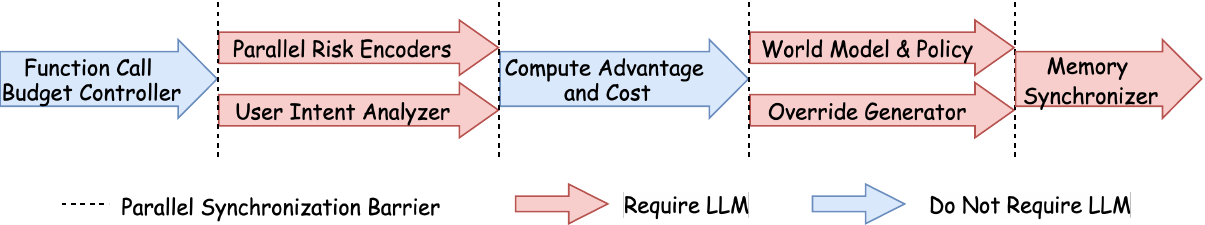}
  \caption{Execution flow of the SafeAgent Core.}
  \label{fig:safeagent_core_graph}
\end{figure*}

\subsection{Core Decision Operators}
\label{subsec:core_operators}

We instantiate the abstract operators introduced in Definition~\ref{defin:core_formalization} through a set of functional components specialized for agent security. These operators jointly enable context-aware risk assessment and intervention under dynamic agent workflows.

\paragraph{Risk Encoding}
The risk encoding operator projects the current contextual fragment into a latent space structured by security-relevant factors. In SafeAgent, this operator is realized as a set of parallel specialist encoders, each capturing a distinct dimension of contextual risk. These specialists act as domain experts over phenomena such as obfuscation, trigger activation, memory poisoning, secret leakage, control hijacking, evasion, policy violation, unsafe tool planning, argument-level risk, and task drift. Their outputs are composed into a unified latent representation that concentrates the security semantics of the current observation while remaining suitable for downstream consequence modeling and intervention.

\paragraph{Advantage--Cost Modeling}
The advantage--cost operator evaluates candidate intervention strategies under the current contextual state. Each candidate, including approval, rejection, or recovery-oriented actions (cf. Section~\ref{subsec:controller_recovery}), is associated with a structured representation of expected utility and operational cost.

This operator does not predict future state evolution, but instead provides an immediate assessment of how each strategy balances task utility against potential risk under the current observation. By modeling the impact of different risk factors across temporal dimensions, it captures how interventions affect system objectives at different stages of execution.

Costs are represented along multiple dimensions reflecting system overhead and performance tradeoffs, enabling consistent comparison across heterogeneous intervention strategies. This unified evaluation framework allows the system to select actions under a coherent safety--utility criterion.

\paragraph{World Model Simulation}
The world model operator performs latent-space simulation of action consequences by leveraging the reasoning capability of the underlying model. Conditioned on the current state and latent representation, it predicts how candidate actions may shape the future trajectory of the system.

This simulation is carried out entirely in the internal representation space, allowing the system to anticipate downstream effects without executing actions in the external environment. In contrast to advantage--cost modeling, which provides immediate evaluation under the current context, the world model captures the dynamic evolution induced by actions across time.

\paragraph{Policy Arbitration}
The policy operator performs configurable decision-making by balancing multiple competing objectives under task-specific requirements. Rather than relying on a fixed decision rule, it adapts to different deployment settings by weighting factors such as task effectiveness, safety, and interaction quality within a unified decision framework.

This arbitration is carried out under explicit constraints: actions that violate critical safety conditions are excluded from consideration, regardless of their potential utility. As a result, the policy enforces a constrained optimization process, selecting admissible interventions that achieve an appropriate balance between utility and risk.

Such a design allows the system to adapt its behavior across heterogeneous tasks while maintaining consistent safety guarantees.

\paragraph{World-State Representation and Memory Synchronization}

The SafeAgent Core maintains an explicit latent world state that captures risk signals across multiple temporal scales. In particular, the state distinguishes between immediate observational risk, task-level risk associated with the ongoing execution process, and long-horizon risk that persists across extended interaction history. This decomposition provides a structured abstraction of how risks arise, propagate, and accumulate over time.

Memory synchronization governs the evolution of this multi-scale state through a measurable transition process. Newly observed signals contribute to the immediate risk representation, which influences task-level risk, and may further accumulate into long-horizon risk. This propagation is not heuristic, but follows a consistent and computable transformation, enabling the system to quantify how risk evolves across temporal scales.

Rather than storing the full interaction history, the state is updated through controlled transformation and compression, ensuring that only decision-relevant risk structure is retained. This enables consistent reasoning over evolving context while preventing uncontrolled growth of the representation.

\subsection{End-to-End Execution Flow}
\label{subsec:e2e_exec_flow}

Figure~\ref{fig:safeagent_core_graph} illustrates the execution flow of the SafeAgent Core as a modular decision pipeline operating over a persistent session state. At each step, the Core receives a normalized observation together with the current state, and produces an intervention decision along with optional structured modifications.

The Core is designed as a configurable and extensible system. It supports pluggable encoding modules that capture different aspects of contextual risk, and allows task-specific policies to be registered prior to session execution. These configurations determine how risk signals are interpreted and how trade-offs between safety and utility are resolved under different deployment scenarios.

During execution, the Core composes its functional operators into a unified decision process. Risk signals are first extracted and transformed into a latent representation, which is then used to evaluate candidate intervention strategies. The system further performs consequence-aware reasoning to anticipate the effects of these strategies, and applies a policy operator to select an admissible action under the given constraints. In parallel, optional override signals may be generated to modify unsafe intermediate content or tool interactions.

All decision outcomes are fed back into the internal state through a synchronized update process, allowing the Core to maintain a coherent view of risk evolution across the session. This stateful execution enables the system to adapt its behavior over time, rather than making isolated decisions based solely on the current observation.

Together, this design defines the SafeAgent Core as a reusable decision module that can be instantiated under different configurations while preserving a consistent execution structure across agentic workflows.

We next evaluate this design empirically. The following section specifies the threat setting, experimental setup, and evaluation metrics, and reports security outcomes under prompt-injection and workflow-level attacks, together with ablation results on recovery and policy design choices.

\section{Evaluation}
\label{sec:evaluation}

\subsection{Experimental Setup}
\label{subsec:experimental_setup}

\paragraph{Agent and Model Setup}
We implement the agent in a ReAct-style framework, using DeepSeek-V3.2 as the underlying backbone for reasoning and tool-use decisions. The SafeAgent Core is instantiated with the gpt-oss-safeguard-20b model~\cite{openai2025gptoss}, which provides the reasoning capability required for risk assessment and intervention.

\paragraph{Benchmarks}
We evaluate SafeAgent on two representative benchmarks for agent security: Agent Security Bench (ASB)~\cite{zhang2025agentsecuritybenchasb} and InjecAgent~\cite{zhan2024injecagentbenchmarkingindirectprompt}. ASB consists of 51 agent tasks combined with 40 attack tools, resulting in 2040 evaluation cases that cover diverse prompt-injection and tool-abuse scenarios in agent workflows. InjecAgent focuses on prompt injection attacks in tool-integrated agents, including 510 direct-harm tasks and 544 data-stealing tasks that model multi-stage attack behaviors such as data extraction and exfiltration.

\paragraph{Attack Setting}
We evaluate SafeAgent under both direct and indirect prompt-injection settings by integrating attacker-controlled tools and inputs into the agent environment. For ASB~\cite{zhang2025agentsecuritybenchasb}, we follow the benchmark protocol and inject adversarial signals through multiple channels. In direct prompt injection (DPI), the attack payload is appended to the end of the user input. In indirect prompt injection (IPI), the payload is injected into external observations, such as tool outputs returned to the agent. In memory poisoning (MP), a malicious assistant message is inserted prior to the user input to contaminate the interaction context before task execution. In all cases, attacker tools are included in the agent’s toolset to enable tool-level attack execution. For InjecAgent~\cite{zhan2024injecagentbenchmarkingindirectprompt}, we use the provided attack datasets to simulate interactions between the agent and the environment. The agent processes adversarial inputs and tool outputs as specified by the benchmark, and its actions are recorded for evaluation.

\paragraph{Defense Methods}
We compare SafeAgent with three defense settings: (i) a baseline agent without additional protection, (ii) Llama Guard~\cite{inan2023llamaguard}, and (iii) LLM Guard~\cite{protectai_llmguard}. 

Llama Guard represents a class of LLM-based safeguards that leverage large language models to evaluate the safety of inputs and outputs, while LLM Guard represents classifier-based approaches built on pretrained models such as RoBERTa for detecting unsafe content. These methods correspond to two widely adopted paradigms for guardrail design in LLM systems. 

In our evaluation, both approaches are applied as text-level defenses over agent interactions. At each step, the current textual context is analyzed, and when a potential threat is detected, the corresponding content is replaced with a sanitized version before execution continues. This ensures that all methods operate under a consistent setting that preserves task continuity while applying defense mechanisms.

\paragraph{Evaluation Metrics}
We evaluate both security and task performance under attack.  Attack success rate (ASR) measures the fraction of cases in which the agent executes an attacker-controlled tool. An attack is considered successful if any attacker tool is invoked during the interaction.

To assess task performance, we use DeepSeek-V3.2 model as a judge to evaluate whether the agent successfully completes the original task. We report the performance under benign conditions as PNA (Performance under No Attack), which reflects the task completion ability without adversarial interference.

To capture the trade-off between safety and utility, we define a combined metric:
\[
\mathrm{NRP} = (1 - \mathrm{ASR}) \times \mathrm{PNA},
\]
which measures normalized reward preservation under attack.

For data-stealing tasks with multi-stage objectives, we further report stage-wise success rates. Specifically, $S1$ measures the success rate of data extraction, and $S2$ measures the success rate of data transmission.

\subsection{Results}
\label{subsec:results}

We compared SafeAgent with baseline and representative guardrail-based defenses on ASB and InjecAgent. Since the two benchmarks differ in task formulation and evaluation protocol, we report their results separately.

\begin{table*}[t]
\centering
\caption{Results on Agent Security Bench (ASB).}
\label{tab:asb_results}
\small
\setlength{\tabcolsep}{5pt}
\renewcommand{\arraystretch}{1.08}
\begin{tabular}{lccccccc}
\toprule
\textbf{Method} &
\textbf{PNA $\uparrow$} &
\textbf{DPI ASR $\downarrow$} &
\textbf{DPI NRP $\uparrow$} &
\textbf{IPI ASR $\downarrow$} &
\textbf{IPI NRP $\uparrow$} &
\textbf{MP ASR $\downarrow$} &
\textbf{MP NRP $\uparrow$} \\
\midrule
No Defense  & \textbf{0.8431} & 0.7627 & 0.2001 & 0.8054 & 0.1641 & 0.4485 & 0.4650 \\
Llama Guard & 0.7843 & 0.5510 & 0.3522 & 0.6858 & 0.2464 & 0.4186 & 0.4560 \\
LLM Guard   & 0.7451 & 0.7588 & 0.1797 & 0.7941 & 0.1534 & 0.4426 & 0.4153 \\
SafeAgent   & 0.7451 & \textbf{0.4186} & \textbf{0.4332} & \textbf{0.2936} & \textbf{0.5263} & \textbf{0.1858} & \textbf{0.6067} \\
\bottomrule
\end{tabular}
\end{table*}

Table~\ref{tab:asb_results} reports the results on ASB. SafeAgent achieves the lowest attack success rates across all three attack categories, including direct prompt injection (DPI), indirect prompt injection (IPI), and memory poisoning (MP). In particular, the reduction is most pronounced under IPI and MP, where SafeAgent lowers ASR to 0.2936 and 0.1858, respectively.

Although the baseline agent without defense attains the highest benign performance (PNA), it is also the most vulnerable to attacks. In contrast, SafeAgent preserves the same benign task performance as LLM Guard while substantially reducing attack success. As a result, it achieves the highest NRP under all three attack types, indicating the best overall balance between attack resistance and task preservation.

The comparison also highlights the limitations of text-level guardrails in agent settings. Llama Guard improves over the undefended baseline, but remains substantially weaker than SafeAgent on all three attack classes. LLM Guard provides only marginal protection against ASB attacks and in some cases remains close to the no-defense setting. These results suggest that message-level filtering alone is insufficient when attacks are realized through workflow state, tool interactions, and multi-step context propagation.

\begin{table*}[t]
\centering
\caption{Results on InjecAgent. S1 denotes data extraction and S2 denotes transmission in data-stealing attacks.}
\label{tab:injecagent_results}
\small
\setlength{\tabcolsep}{7pt}
\renewcommand{\arraystretch}{1.08}
\begin{tabular}{lcccc}
\toprule
\textbf{Method} &
\textbf{Direct Harm $\downarrow$} &
\textbf{S1 $\downarrow$} &
\textbf{S2 $\downarrow$} &
\textbf{Total $\downarrow$} \\
\midrule
No Defense  & 0.8608 & 0.9338 & 0.9921 & 0.9265 \\
Llama Guard & 0.4314 & 0.2279 & 0.7419 & 0.1691 \\
LLM Guard   & 0.8431 & 0.9301 & 0.9308 & 0.8658 \\
SafeAgent   & \textbf{0.3157} & \textbf{0.0000} & \textbf{0.0000} & \textbf{0.0000} \\
\bottomrule
\end{tabular}
\end{table*}

Table~\ref{tab:injecagent_results} shows the results on InjecAgent. SafeAgent achieves the lowest attack success rate on direct-harm tasks and completely suppresses the two-stage data-stealing attacks, reducing both $S1$ and $S2$ to zero. Consequently, its overall attack success rate on the benchmark is also reduced to zero.

This result is particularly important because InjecAgent evaluates attacks that unfold across multiple stages rather than through a single unsafe output. In such settings, the security challenge lies not only in detecting a suspicious fragment, but in preventing the agent from progressing along an unsafe execution trajectory. The results therefore support the claim that SafeAgent’s stateful runtime control is effective against staged attacks that require both intermediate compliance and downstream action execution.

The contrast with existing guardrail methods is again clear. Llama Guard reduces attack success in comparison with the undefended baseline, especially on the first stage of data-stealing attacks, but still allows substantial leakage in the second stage. LLM Guard remains close to the undefended baseline across all InjecAgent metrics. This suggests that purely text-centered safeguards struggle to prevent attacks once malicious intent is carried through tool-mediated interactions and accumulated session context.

Overall, the benchmark results show that SafeAgent consistently provides stronger protection than baseline and guardrail-based alternatives, while maintaining competitive benign-task performance. In the following subsection, we further analyze the contribution of individual design choices through ablation studies.

\subsection{Ablation Study}
\label{subsec:ablation}

\begin{table*}[t]
\centering
\caption{Effect of override confidence on ASB.}
\label{tab:override_confidence}
\small
\setlength{\tabcolsep}{5pt}
\renewcommand{\arraystretch}{1.08}
\begin{tabular}{lccccccc}
\toprule
\textbf{Override Confidence} &
\textbf{PNA $\uparrow$} &
\textbf{DPI ASR $\downarrow$} &
\textbf{DPI NRP $\uparrow$} &
\textbf{IPI ASR $\downarrow$} &
\textbf{IPI NRP $\uparrow$} &
\textbf{MP ASR $\downarrow$} &
\textbf{MP NRP $\uparrow$} \\
\midrule
0.0 & \textbf{0.7843} & 0.4740 & 0.4125 & 0.3191 & 0.5340 & \textbf{0.1750} & \textbf{0.6470} \\
0.5 & 0.7451 & 0.4186 & 0.4332 & 0.2936 & 0.5263 & 0.1858 & 0.6066 \\
1.0 & 0.7451 & \textbf{0.3701} & \textbf{0.4693} & \textbf{0.2733} & \textbf{0.5415} & 0.1779 & 0.6125 \\
\bottomrule
\end{tabular}
\end{table*}

\begin{table*}[t]
\centering
\caption{Effect of policy weighting on ASB.}
\label{tab:policy_ablation}
\small
\setlength{\tabcolsep}{5pt}
\renewcommand{\arraystretch}{1.08}
\begin{tabular}{lccccccc}
\toprule
\textbf{Policy} &
\textbf{PNA $\uparrow$} &
\textbf{DPI ASR $\downarrow$} &
\textbf{DPI NRP $\uparrow$} &
\textbf{IPI ASR $\downarrow$} &
\textbf{IPI NRP $\uparrow$} &
\textbf{MP ASR $\downarrow$} &
\textbf{MP NRP $\uparrow$} \\
\midrule
Safety-first policy & 0.5490 & \textbf{0.2114} & 0.4329 & \textbf{0.1779} & 0.4513 & \textbf{0.1270} & 0.4793 \\
Task-first policy   & \textbf{0.7451} & 0.4186 & \textbf{0.4332} & 0.2936 & \textbf{0.5263} & 0.1858 & \textbf{0.6066} \\
\bottomrule
\end{tabular}
\end{table*}

\paragraph{Override confidence}
We ablate the \emph{override confidence} parameter, which controls how strongly SafeAgent trusts context repair relative to explicit rejection. A lower value corresponds to a conservative regime in which rewritten context is treated as unreliable, while a higher value shifts the policy toward repair-oriented recovery.

Table~\ref{tab:override_confidence} shows that override confidence directly changes how SafeAgent interprets the role of recovery. At confidence 0.0, the system relies less on rewriting and behaves more conservatively, which yields the highest benign performance and the best result under memory poisoning. This suggests that when contamination is persistent and structurally embedded in the interaction history, explicit rejection remains preferable to optimistic repair.

As override confidence increases, SafeAgent becomes more willing to preserve execution by repairing unsafe context. This consistently improves DPI and IPI robustness, with the best ASR and NRP obtained at 1.0, indicating that repair-oriented intervention is effective when attacks are primarily localized within the current contextual fragment. Taken together, these results show that repair is not universally beneficial: its value depends on whether the attack is locally repairable or temporally persistent.

\paragraph{Policy Weight Ablation}
We also compare two policy configurations, safety-first policy and task-first policy, to examine how objective weighting affects the final intervention decision. The former assigns greater emphasis to risk control, whereas the latter prioritizes task completion while retaining a weaker safety preference.

Table~\ref{tab:policy_ablation} reveals a clear safety--utility trade-off induced by policy weighting. The safety-first policy consistently reduces ASR across all attack types, confirming that stronger preference for risk control leads to more conservative interventions. However, this comes with a substantial drop in PNA, indicating that the system suppresses not only unsafe trajectories but also a significant portion of benign task execution.

By contrast, task-first policy preserves substantially more benign utility and therefore achieves higher NRP across DPI, IPI, and MP, despite allowing higher attack success rates. The implication is that policy arbitration does not merely fine-tune behavior, but determines where SafeAgent operates within the safety--utility space. In this sense, the policy module functions as the primary mechanism that turns risk signals into deployment-dependent intervention behavior.

Overall, the ablations show that SafeAgent’s effectiveness depends not only on detecting risk, but on how recovery and policy preferences are configured.

\section{Conclusion and Future Work}
\label{sec:conclusion}

In this work, we presented SafeAgent, a runtime security architecture for agentic systems that combines a governed execution controller with an external context-aware decision core. The central motivation of our design is that securing LLM-based agents requires mechanisms that operate throughout execution, rather than relying solely on input-output filtering. By separating execution control from semantic risk reasoning and maintaining a persistent session state, SafeAgent provides a flexible and configurable framework for workflow-level protection.

Our results show that this architectural design is effective against prompt-injection and workflow-level attacks in tool-using agents. Beyond a practical implementation, SafeAgent also provides a general decision framework for agent security, in which safety is treated as a stateful and context-dependent intervention problem. In addition, the modular organization of the Core supports parallel execution of selected stages, which may improve deployability in practice.

At the same time, the proposed approach still incurs non-trivial computational overhead, which remains an important limitation of the current system. Future work will focus on extending SafeAgent as a more general protection platform, including finer-grained control over tool capabilities and more flexible security-policy configuration. We also plan to improve efficiency by reducing latency in the internal decision loop, and to explore whether selected components of the system can benefit from learning-based optimization, including reinforcement learning in suitable agentic settings. Finally, although SafeAgent improves robustness against prompt injection, the problem is not fully solved, and emerging threat classes such as skill injection require further study.

\bibliographystyle{IEEEtran}
\bibliography{journal}

\end{document}